\documentclass[letterpaper]{article} 
\usepackage{aaai24}  
\usepackage{times}  
\usepackage{helvet}  
\usepackage{courier}  
\usepackage[hyphens]{url}  
\usepackage{graphicx} 
\usepackage{natbib}  
\usepackage{caption} 
\usepackage{algorithm}
\usepackage{algorithmic}
\usepackage[ruled,vlined,algo2e]{algorithm2e}
\usepackage{tabularx} 
\usepackage{multirow}
\usepackage{multicol}

\usepackage{subfigure}
\usepackage{amsmath,amssymb,amsfonts}
\usepackage{cite}
\usepackage{textcomp}
\usepackage{subfigure}
\usepackage{xcolor}

\usepackage{newfloat}
\usepackage{listings}
\DeclareCaptionStyle{ruled}{labelfont=normalfont,labelsep=colon,strut=off} 
\floatstyle{ruled}
\newfloat{listing}{tb}{lst}{}
\floatname{listing}{Listing}
\title{Confusing Pair Correction Based on Category Prototype for Domain Adaptation under Noisy Environments}
\author {
    Churan Zhi\textsuperscript{\rm 1},
    Junbao Zhuo\textsuperscript{\rm 2}\thanks{Corresponding author.},
    Shuhui Wang\textsuperscript{\rm 2}
}
\affiliations {
    \textsuperscript{\rm 1}Beijing Jiaotong University, Beijing, China\\
    \textsuperscript{\rm 2}Institute of Computing Technology, Chinese Academy of Sciences, Beijing, China\\
    leo.churanzhi@hotmail.com, junbao.zhuo@vipl.ict.ac.cn,
    wangshuhui@ict.ac.cn
}

\usepackage{bibentry}

\begin{document}

\maketitle

\begin{abstract}
In this paper, we address unsupervised domain adaptation under noisy environments, which is more challenging and practical than traditional domain adaptation. In this scenario, the model is prone to overfitting noisy labels, resulting in a more pronounced domain shift and a notable decline in the overall model performance. Previous methods employed prototype methods for domain adaptation on robust feature spaces. However, these approaches struggle to effectively classify classes with similar features under noisy environments. To address this issue, we propose a new method to detect and correct confusing class pair. We first divide classes into easy and hard classes based on the small loss criterion. We then leverage the top-2 predictions for each sample after aligning the source and target domain to find the confusing pair in the hard classes. We apply label correction to the noisy samples within the confusing pair. With the proposed label correction method, we can train our model with more accurate labels. Extensive experiments confirm the effectiveness of our method and demonstrate its favorable performance compared with existing state-of-the-art methods. Our codes are publicly available at https://github.com/Hehxcf/CPC/.
\end{abstract}

\section{Introduction}
Domain shift between train and test data usually leads to a deterioration in the models' performance. Therefore, numerous methods~\cite{b18, b19, b22} have been proposed to mitigate the domain shifts. However, these studies often assume that the source labels are clean. In reality, obtaining clean datasets is time-consuming and expensive. In many cases, it is simpler to acquire weakly labeled data via the social media. However, such data inevitably contain noise, leading to a more challenging situation, named unsupervised domain adaptation (UDA) under noisy environments.

The main challenge of UDA under noisy environments is that the model tends to fit noisy labels, leading to adverse effects on the overall training process. Moreover, due to label noise, it is more difficult to classify two classes that have similar features, posing additional difficulties in the domain adaptation process. Therefore, the key to addressing UDA under noisy environments is to correct noisy labels and clarify classification boundaries for similar classes.

Some existing methods~\cite{tcl,rda,csr} for UDA under noisy environments use small loss criteria to separate clean and noisy samples and obtain pseudo labels from the label space. However, existing works~\cite{can} empirically show that using information from the feature space is more robust. For instance, the feature-based method CAN~\cite{can}, outperforms the label-based algorithms, TCL~\cite{tcl} and RDA~\cite{rda}, on the office-31 dataset with 40\% label corruption.

\begin{figure}
\centering
\includegraphics[width=\columnwidth]{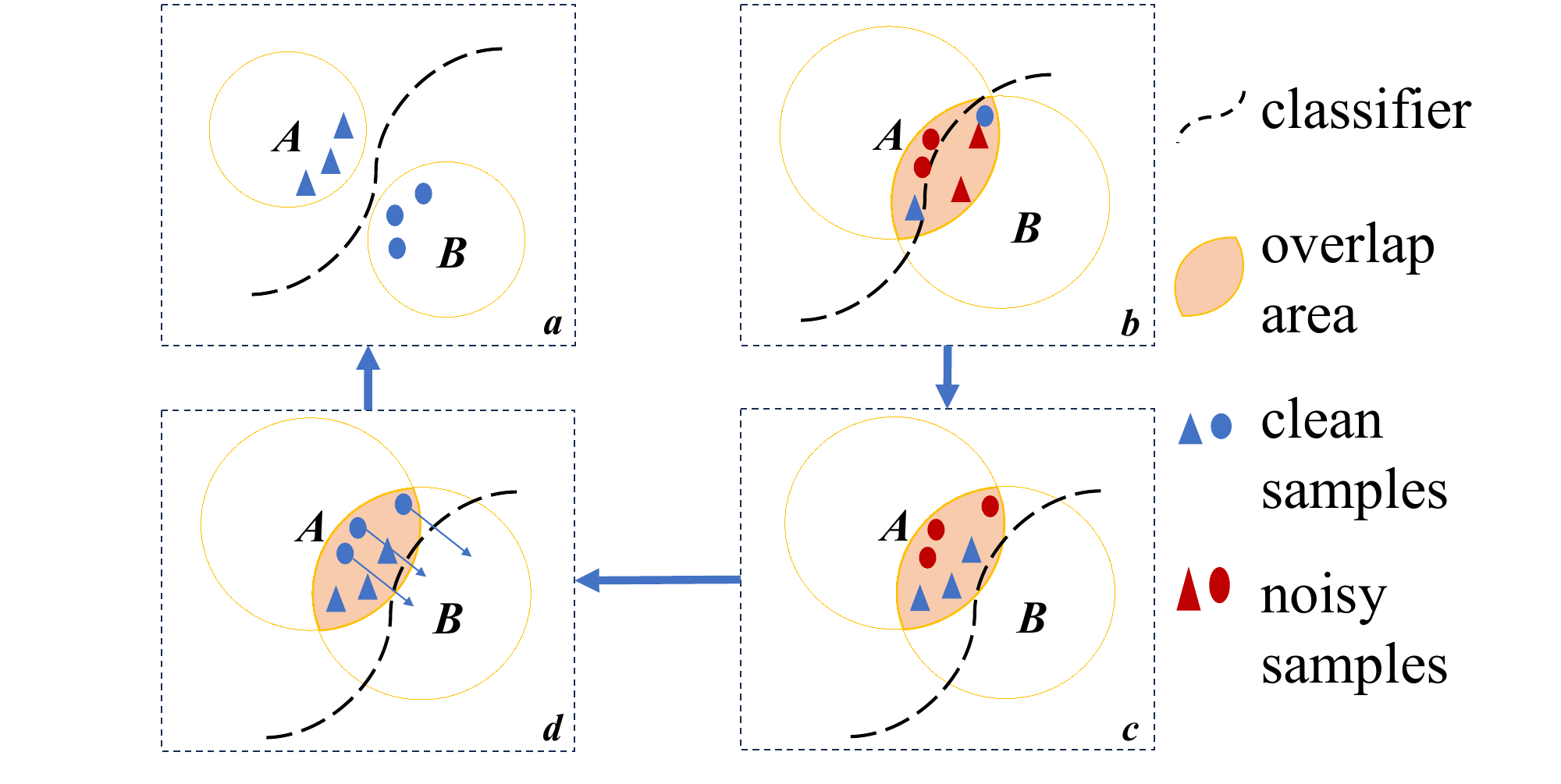}
\caption{Demonstration of our confusing pair correction strategy. (b) Noisy samples of the confusing class pair frequently switch their predicted labels. (c) Class A becomes dominant, class B becomes weak. Samples of B will be misclassified as A. (d) Label correction between the confusing class pair. (a) The classification boundary between the two classes becomes clearer.}
\vspace{-2ex}
\label{confusionshow}
\end{figure}

However, those feature-based methods overlook the phenomenon that two classes with similar features are more likely to overlap in the feature space, which becomes more evident in the presence of label noise. The predictions of samples near the classification boundary of the two classes switch frequently (Figure~\ref{confusionshow} (b)). As the model converges, one class will dominate, and samples of the other class will be misclassified as the dominant class (Figure~\ref{confusionshow} (c)), which severely deteriorates the model's performance.

To address the issue mentioned above, we propose a novel method that detects the confusing class pair, two classes that have similar features. We apply the confusing pair correction strategy in the confusing zone, the overlap area between the two classes (Figure~\ref{confusionshow} (d)). With this method, the pairs are less likely to overlap, resulting in a more distinct classification boundary between them (Figure \ref{confusionshow} (a)). 

Specifically, we establish a shared prototype to identify confusing class pair according to the sample distribution among two classes. The shared prototype is constructed with the noisy labeled source domain and the pseudo-labeled target domain for mitigating domain shift and resulting in more accurate pseudo labels of the target domain. Subsequently, we remove the farthest samples from the prototype to enhance the clarity of classification boundaries. More accurate pseudo labels of the target domain help capture a more accurate sample distribution of classes.

We further categorize classes into easy and hard groups by utilizing the small loss criterion, as we assume that the confusing class pair appear in these hard classes. 
We leverage the top-2 (class pair) predictions of target samples to capture the confusing pair between the two classes. The frequency of a specific class pair indicates the probability that the rank 1 predicted class to be misclassified into the rank 2 predicted class. We then compute the frequencies for all pairs and subsequently identify the confusing class pair $\alpha^*$ and $\beta^*$ for label correction. Specifically, the labels of samples with low confidence from class $\alpha^*$ are corrected into $\beta^*$. Therefore, We are able to train our model with corrected pseudo labels and achieve better performance.


We conduct extensive experiments on three widely used benchmarks to demonstrate the effectiveness of our method. We achieve the accuracy of  64.7\%, 87.4\%, and 85.5\% under 40\% label corruption in Office-home, Office-31, and Bing-Caltech datasets, respectively. The key contributions are summarized as follows.
\begin{itemize}
\item 
We present an innovative confusing pair correction method based on the prototype. We calculate the frequency of the top-2 predictions for target samples to identify the confusing class pair. We then perform label correction on noisy samples of confusing pairs to obtain more accurate pseudo labels for training a better model.
\item 
Extensive experimental results demonstrate that our method surpasses state-of-the-art approaches or achieves comparable performance.

\end{itemize}

\section{Related Work}
\textbf{Unsupervised domain adaptation (UDA)} is to transfer knowledge from a source domain with plenty of labels to an unlabeled target domain~\cite{b20, b21, afs}. Recently, many methods have been proposed to address UDA by reducing domain shift. These methods can be mainly categorized into two groups: divergence-based and adversarial-based methods.

Divergence-based methods ~\cite{uodr} mainly focus on minimizing the distribution discrepancy across two domains. Deep CORAL~\cite{b25} aligns the second-order statistics of the source and target distributions with a linear transformation. CAN~\cite{can} and DDC~\cite{b26} aim to measure and minimize domain discrepancy with the maximum mean discrepancy. Some other methods~\cite{b27, b30} utilize certain distance metrics to reduce domain discrepancy. TPN~\cite{b24} jointly bridges the domain gap by minimizing multi-granular domain discrepancies and building classifiers with noisy labeled source data and unlabeled target data. Adversarial-based methods~\cite{b34, b35} get inspiration from Generative Adversarial Networks (GANs)~\cite{b28}. The objective is to acquire domain-invariant features through adversarial learning. DANN~\cite{dann} uses a discriminator to distinguish source and target features and a feature extractor to fool them. Based on the DANN, ADDA~\cite{adda} and MCD~\cite{b29} extend this architecture to incorporate multiple feature extractors and classifiers.

Class Confusion is a common phenomenon encountered during domain adaptation. If two classes have very similar features, models tend to confuse them. To handle the confusion, ALDA~\cite{b31} employs a discriminator network to produce a confusion matrix to correct the noisy pseudo labels. MCC~\cite{b23} minimizes the cross-class confusion matrix to mitigate the class confusion effects. However, all these methods mentioned above assume the source domain is clean. In reality, the source domain contains noise, and the model tends to overfit noisy labels, further exacerbating the domain shift. Furthermore, the classification accuracy of similar classes can deteriorate as well.

\textbf{Domain adaptation under noisy environments} is more practical and challenging than unsupervised domain adaptation. Solving such problems involves three steps: identifying clean samples from lots of noisy samples, selecting and correcting noisy labels, and using clean samples for training. 
For example, TCL~\cite{tcl} trains a model with a transferable curriculum that identifies clean and transferable source samples. RDA~\cite{rda} employs an offline curriculum learning approach to choose clean samples and mitigates the effect of feature noise by minimizing a margin discrepancy based on a proxy distribution.
CSR~\cite{csr} leverages the outputs of two symmetrical domain adaptation models and considers both consistent and inconsistent samples. However, CSR is very complex because it relies on two domain adaptation models. While the mentioned methods above achieved certain accuracy, they did not address the issue of confusion between two similar classes. Our approach addresses this issue by confusing class pair correction.


\begin{figure*}[h]
    \begin{center}
    \includegraphics[width=\textwidth]{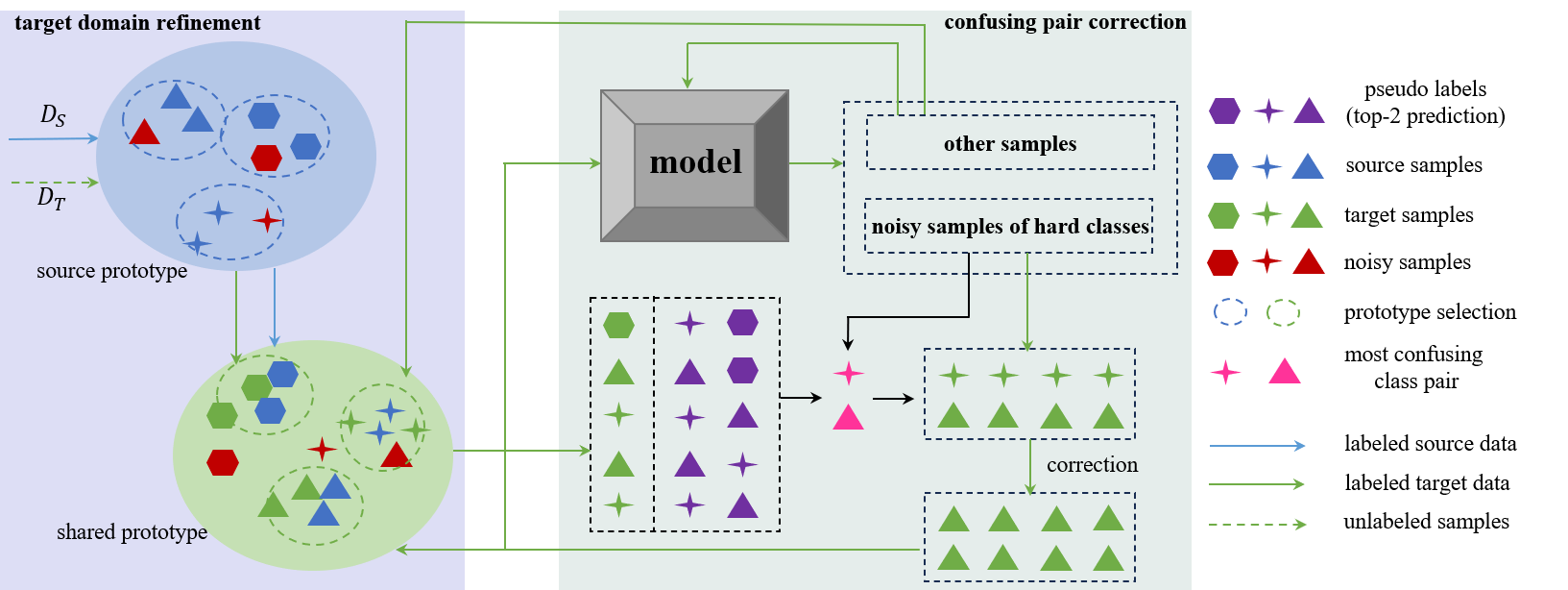}
    \caption{The framework of our method. It contains two components: target domain refinement and confusing pair correction. Firstly, we construct a shared prototype to align the source and target domains and obtain pseudo labels for the target samples. Next, pair labels are generated from the prototype by using top-2 prediction. Secondly, the confusing pair correction method finds the most confusing pair and performs label correction for the noisy samples of the pair. Once the labels are corrected, the prototypes and all pseudo labels are recalculated.}
    \vspace{-2ex}
    \label{fig:2}
    \end{center}
\end{figure*}
\section{Methodology}
Unsupervised domain adaptation (UDA) seeks to transfer knowledge from a labeled source domain to an unlabeled discrepant target domain. Formally, the source and target domains are defined as $\mathcal{D}_s=\left\{\left(x_i^s,y_i^s\right)\right\}_{i=1}^{N_{s}}$, $\mathcal{D}_t=\left\{x_i^t\right\}_{i=1}^{N_{t}}$, and $x_i^s, x_i^t \in \mathbb R^d$ and $y_i^s \in \left\{1,..., C\right\}$, where $N_{s}$ $(N_{t})$ is the total number of source (target) instances, $d$ is the feature dimension and $C$ is the number of classes. In real-world scenarios, it is simpler and time-saving to obtain weakly labeled data via the internet. Therefore, the labels of the source domain often contain noise. Specifically, it is usually assumed that $p_{noise}$ of each class will be randomly flipped to another class with equal probability. In this case, models potentially learn noisy knowledge, which hinders their ability to achieve accurate predictions.

Detecting noise labels and correcting them is crucial for UDA under noisy environments. To address this issue, we propose to fully utilize the top-2 predictions of a model in the robust feature space and perform confusing pair detection and correction subsequently. The proposed architecture for the learning process is shown in Figure \ref{fig:2}.

\subsection{Target Domain Refinement}
In our approach, we first construct a prototype using the source domain and then predict pseudo labels for the target domain with the prototype. Then, the prototype for the target domain is generated using pseudo labels and aligned with the source prototype. This process does not require training and can be iterated. 

Given the source domain $\mathcal{D}_s=\left\{\left(x_i^s,y_i^s\right)\right\}_{i=1}^{N_{s}}$, the prototype of class $k$ for the source domain $P_k^s$ of shape $C\times d$, is computed as follows:
\begin{equation}
{P}_{k}^{s}=\frac{1}{\left|{\mathcal D}_{k}^{s}\right|}\sum_{{x}_{i}^{s}\in{{D}_{k}^{s}}}\phi\left({x}_{i}^{s}, \theta\right),
\end{equation}
where ${P}_{k}^{s}$ is the prototype, the average of all embedded samples of class ${k}$ from the source domain, and $\phi$ is the feature extractor of the model.

After obtaining the prototype for the source domain, we initialize the shared prototype $P$ as $P^s$ and obtain the pseudo label of a target sample ${x}_{i}^{t}$ via cosine similarity:
\begin{equation}
{{\hat{y}}_i^{t}} = \mathop{\arg\max} \left(\phi\left({x}_{i}^{t}, \theta\right)*P^\mathrm{T}\right),
\end{equation}
where $P^\mathrm{T}$ is the transpose of $P$, $y_i^t$ is
the i-th sample of the target domain.

Similarly, the prototype for category $k$ of target domain $P_k^t$ is computed as follows:
\begin{equation}
{P}_{k}^{t}=\frac{1}{\left|{\mathcal D}_{k}^{t}\right|}\sum_{{x}_{i}^{t}\in{{D}_{k}^{t}}}\phi\left({x}_{i}^{t}, \theta\right).
\end{equation}

We add $P_k^t$ to $P$ as $P=(P+P_k^t)/2$, to align the source and target domains in the high-dimensional space.

Note that both source and target domains contain noise. In this scenario, samples that are farthest from the prototype of each class are considered to be noise, which degrades classification boundaries. Therefore, we discard these samples and exclude them from the prototype calculation. The distance between a sample $x_i^t$ and the prototype ${P}_{k}$ of category $k$ is computed by Euclidean distance:

\begin{equation}
dist\left(x_i^{t}, P_k\right)={||x_i^{t},P_k||}_2.
\end{equation}

We select $\tau$ of the farthest samples of each class computed by Eq. (5) and discard them, where $\tau$ is a ratio ranging from 0 to 1. Formally, we sort samples of each class in ascending order based on their distances to the corresponding prototype.
and obtain$x_{1}^{t}, x_{2}^{t},...,x_{\lfloor\tau*n_{kt}\rfloor}^{t},...,x_{n_{kt}}^{t}$ for the target samples, where $n_{kt}$ denotes the number of target samples in class $k$. Thus, the de-noised prototype is re-calculated by:

\begin{equation}
Q_t\left({P}_{k}^{t}\right)=\frac{1}{\lfloor\tau*n_{kt}\rfloor}\sum_{i=1}^{{\lfloor\tau*n_{kt}\rfloor}}\phi\left({x}_{i}^{t}, \theta\right).
\end{equation}
The de-noised prototype $Q_t(P_k^t)$ is adopted to iteratively refine the shared prototypes $P$. Therefore, the prototype $P_{k,i}$ of class k for the i-th iteration can be formulated as:
\begin{equation}
{P}_{k,i} = \frac{{P}_{k,i-1} + Q_t\left({P}_{k,i}^{t}\right)}{2}.
\end{equation}
\begin{algorithm}[H]
\caption{The training procedures of our method.}
\label{algorithm}
\SetAlgoLined			
\DontPrintSemicolon		
\KwIn{Noisy labeled source images $\mathcal{D}_s=\left\{\left(x_i^s,y_i^s\right)\right\}_{i=1}^{N_{s}}$; Target images ${\mathcal D}_{t}=\left\{{x}_{i}^{t}\right\}_{i=1}^{N_{t}}$; Initial feature extractor ${\mathit{\Phi}}$}
\For {$correction\;epoch\;s = 1, 2... S$}{
    \For {$train\;epoch\;e = 1,2,...,E$}{
        \For {$refinememt\;epoch\;m = 1,2,...,M$}{
            // \textbf{Target Domain Refinement.}\\
            Compute $P^s$ from source by Eq. (1);\\
            Generate pseudo labels $\left\{\hat{y}_i^t\right\}_{i=1}^{N_{t}}$ by Eq. (2);\\
            Compute $P^t$, update $P$ by Eq. (3);\\
            Re-calculate the de-noised prototype by Eq. (5);\\
            Iterate Prototype by Eq. (6).
        }
        Use cross-entropy loss to train model by using $\mathcal{D}_t=\left\{\left(x_i^t, {\hat{y}}_i^{t}\right)\right\}_{i=1}^{N_{t}}$ obtained in Eq. (2).\\
    }
    // \textbf{Confusing Pair Correction.} \\
    Divide samples into clean and noisy samples. Compute set $S_{hc}$ by Eq. (7);\\
    Generate pair labels for each target sample using the top-2 predictions through Eq. (8). \\
    Compute ${\alpha}^{*}, {\beta}^{*}$ by Eq. (9), (10).
    Do the most confusing pair correction.
}
return ${\mathit{\Phi}}$
\end{algorithm}
where $P_{k, i}$ is the shared prototype of class k for iteration $i-1$ and ${P}_{k, i}^{t}$ is the prototype of class k of the target domain for iteration $i$. After obtaining the prototype $P$, we use Eq. (2) to refine the pseudo labels for the target domain. The resulting refined target domain can be defined as: $\mathcal{D}_t=\left\{\left(x_i^t, {\hat{y}}_i^{t}\right)\right\}_{i=1}^{N_{t}}$.

\subsection{Confusing Pair Correction}
After applying target domain refinement, it is important to note that there exists noise in the refined target domain. Moreover, refining the pseudo labels via prototype tends to confuse two classes with similar features. To address these issues, we propose to identify the confusing class pair and correct the corresponding wrong pseudo labels.

We assume that there exist hard classes, in which the confusing class pair may appear in these hard classes. 
According to the small loss criterion, the average losses of samples from the hard class are assumed to be much larger than the average losses of samples from the easy class. Based on the above assumptions, we obtain hard classes set by Eq. (7):
\begin{equation}
S_{hc}=\left\{k \Bigg| \frac{1}{\left|{\mathcal D}_{k}^{t}\right|}\sum_{{\hat{y}}_i^t=k}L\left({x}_{i}^{t}\right) \leq \zeta\right\},
\end{equation}
where $L\left(\cdot\right)$ denotes the cross entropy loss of a certain sample, $S_{hc}$ denotes the set of hard classes. $k$ is the index of the class. $\zeta$ is a threshold.

To further identify the confusing class pair, we gather the top-2 predictions for every target sample. The top-2 prediction matrix $M_k\in\mathbb{R}^{N_t\times2}$ can be represented as:

\begin{equation}
M_k=\left[\begin{array}{cc}
{\hat{y}}_{11}^t & {\hat{y}}_{12}^t \\
{\hat{y}}_{21}^t & {\hat{y}}_{22}^t \\
\vdots & \vdots \\
{\hat{y}}_{{N_t}1}^t & {\hat{y}}_{{N_t}2}^t
\end{array}\right],
\end{equation}
where ${\hat{y}}_{i1}^t, {\hat{y}}_{i2}^t$ denote the most confident (rank 1) prediction and the second confident (rank 2) prediction of the i-th sample, and ${\hat{y}}_{i1}^t$ have more vote counts than ${\hat{y}}_{i2}^t$.

The most confusing pair $\alpha^*, \beta^*$ should appear very frequently in the matrix $M_k$. 
Therefore, we first compute the frequency of a certain pair label $\alpha, \beta$ in $M_k$:
\begin{equation}
f^{\alpha,\beta}=\frac{\sum\mathbb{I}\left( y_{\alpha}==M_k\left[i\right]\left[1\right]\land y_{\beta}==M_k\left[i\right]\left[2\right]\right)}{\sum\mathbb{I}\left(  y_{\alpha}==M_k\left[i\right]\left[1\right]\right)},
\end{equation}
where $\mathbb{I}$ is the indicator function, $f^{\alpha,\beta}$ denotes the probability that class $\alpha$ tends to flip into class $\beta$.
Thus, $f^{\alpha,\beta} > f^{\beta,\alpha}$ indicates that class $\alpha$ has a higher likelihood of flipping into class $\beta$, while class $\beta$ is not as easily confused with class $\alpha$.


For simplicity, we only choose the most confusing pair $\alpha^*, \beta^*$ for label correction. $\alpha^*, \beta^*$  can be obtained via Eq. (10):
\begin{equation}
\begin{aligned}
\alpha^*,\beta^*=\mathop{\arg\max}_{\alpha,\beta} f^{\alpha,\beta} \land \left(f^{\alpha,\beta}>f^{\beta,\alpha}\right)\\
\land\left(\alpha,\beta\in S_{hc}\right).
\end{aligned}
\end{equation}



Recall that \textbf{noisy samples} in class $\alpha$ have a higher likelihood of flipping into class $\beta$, but not otherwise. The noisy samples are selected by the small loss criterion. Therefore, we then correct the labels of noisy samples whose pseudo labels are $\alpha^*$ to $\beta^*$.


Once we correct the pseudo labels, we adopt cross-entropy loss with these pseudo labels to train the model. The training process is illustrated in Algorithm (1).

\begin{table*}[h]
\small
\centering
\setlength{\tabcolsep}{1.3mm}{
\scalebox{1}{%
\begin{tabular}{cccccccccccccc}
\hline
Method
&Ar$\rightarrow$Cl &Ar$\rightarrow$Pr &Ar$\rightarrow$Rw
&Cl$\rightarrow$Ar &Cl$\rightarrow$Pr &Cl$\rightarrow$Rw
&Pr$\rightarrow$Ar &Pr$\rightarrow$Cl &Pr$\rightarrow$Rw
&Rw$\rightarrow$Ar &Rw$\rightarrow$Cl &Rw$\rightarrow$Pr
&Avg.\\
\hline
ResNet&19.8&37.8&46.5&22.3&32.1&30.5&20.5&13.3&37.0&31.8&19.8&50.1&30.1 \\
DANN&25.3&40.4&51.9&36.5&43.2&48.3&34.7&25.8&54.6&46.2&34.3&61.3&41.9 \\
MDD&42.2&59.9&66.9&47.2&59.0&59.8&40.6&34.5&60.9&55.2&42.9&73.3&53.5 \\
TCL&21.1&35.6&61.4&16.1&44.6&36.4&24.6&30.4&68.7&59.9&25.7&68.6&44.1 \\
RDA&40.3&56.9&64.3&46.9&57.1&59.7&41.2&32.6&59.7&51.1&42.0&71.0&51.9 \\
CAN&27.7&53.1&59.5&33.1&55.8&53.2&31.3&30.3&56.6&38.4&33.6&65.6&44.9 \\
CSR (CAN)&40.2&62.4&65.3&55.4&67.5&63.8&52.3&\textbf{47.2}&68.7&64.3&\textbf{51.0}&73.5&59.3 \\

\hline
\hline
CPC&\underline{41.6}&\underline{68.2}&\underline{75.6}&\textbf{63.3}&\textbf{72.4}&\underline{72.4}&\underline{60.9}&43.4&\underline{75.6}&\underline{65.6}&46.7&\underline{76.0}&\underline{63.5} \\
CPC (CAN)&\textbf{42.6}&\textbf{70.8}&\textbf{76.8}&\underline{63.0}&\underline{72.3}&\textbf{73.5}&\textbf{61.7}&\underline{44.3}&\textbf{78.1}&\textbf{66.3}&\underline{47.4}&\textbf{79.0}&\textbf{64.7}\\
\hline
\end{tabular}}}
\caption{Classification accuracy (\%) of Office-Home under 40\% label corruption. The best results are marked in bold font, and the second-best results are underlined.}

\label{result65}
\end{table*}

\begin{table*}[h]
\centering
\setlength{\tabcolsep}{1.8mm}{
\scalebox{1}{%
\begin{tabularx}{\textwidth}{XXXXXXXX}
\hline
Method
&A$\rightarrow$W &W$\rightarrow$A &A$\rightarrow$D
&D$\rightarrow$A &W$\rightarrow$D &D$\rightarrow$W
&Avg.\\
\hline
ResNet&47.2&33.0&47.1&31.0&68.0&58.8&47.5 \\
SPL&72.6&50.0&75.3&38.9&83.3&64.6&64.1 \\
MentorNet&74.4&54.2&75.0&43.2&85.9&70.6&67.2 \\
DAN&63.2&39.0&58.0&36.7&71.6&61.6&55.0 \\
RTN&64.6&56.2&76.1&49.0&82.7&71.7&66.7 \\
DANN&61.2&46.2&57.4&42.4&74.5&62.0&57.3 \\
ADDA&61.5&49.2&61.2&45.5&74.7&65.1&59.5 \\
MDD&74.7&55.1&76.7&54.3&89.2&81.6&71.9 \\
TCL&82.0&65.7&83.3&60.5&90.8&77.2&76.6 \\
RDA&89.7&67.2&92.0&65.5&96.0&92.7&83.6 \\
CAN&86.7&71.5&90.0&73.7&93.4&90.6&84.3\\
CSR (CAN)&\textbf{91.6}&71.7&\textbf{93.0}&74.8&95.2&94.7&86.8 \\
UMRDA&\underline{90.5}&\textbf{75.0}&90.0&\underline{75.0}&\underline{96.5}&95.5&\underline{87.1} \\
\hline\hline
CPC&87.9&\underline{73.0}&\underline{92.2}&74.2&94.9&\underline{95.5}&86.3 \\
CPC (CAN)&88.1&\textbf{75.0}&91.3&\textbf{77.0}&\textbf{96.7}&\textbf{96.0}&\textbf{87.4} \\
\hline
\end{tabularx}}}
\caption{Classification accuracy (\%) of Office-31 under 40\% label corruption. The best results are marked in bold font, and the second-best results are underlined.}

\label{result31}
\end{table*}
\begin{table*}[h]
\centering
\begin{tabular}{ccccccccccccc}
\hline
Method&ResNet&SPL&MentorNet&RTN&MDD&TCL&RDA&CAN&CSR (CAN)&UMRDA&CPC (CAN) \\
\hline
Acc.&74.4&75.3&75.6&75.8&78.9&79.0&81.7&78.0&{84.4}&\textbf{85.8}&\underline{85.5} \\
\hline

\end{tabular}
\caption{Classification accuracy (\%) achieved on Bing-Caltech with natural corruption. The best results are marked in bold font, and the second-best results are underlined.}

\label{resultbc}
\end{table*}
\begin{table*}
\centering%
\begin{tabularx}{\textwidth}{XXXXXXXXXXX}
\hline
Method &TR-1.0 &TR-0.8 &CPC &A$\rightarrow$W &W$\rightarrow$A &A$\rightarrow$D
&D$\rightarrow$A &W$\rightarrow$D &D$\rightarrow$W
&Avg.\\
\hline
\hline
CPC &$\checkmark$ &$\times$ &$\times$ &83.7&69.9&86.1&72.4&93.5&94.2 &83.3\\
CPC &$\times$ &$\checkmark$ &$\times$ &84.5&70.8&87.1&73.2&92.9&93.5&83.6 \\
CPC &$\times$ &$\checkmark$ &$\checkmark$ &\textbf{87.9}&\textbf{73.0}&\textbf{92.2}&\textbf{74.2}&\textbf{94.9}&\textbf{95.5}&\textbf{86.3} \\
\hline
\end{tabularx}
\caption{Ablation study on Office-31 under 40\% label corruption. TR-1.0, and TR-0.8 denote the target domain refinement with $\tau=1.0, 0.8$, respectively. CPC denotes the most confusing pair correction. The best results are marked in bold font.}
\label{resultaba}

\end{table*}
\section{Experiments}
\subsection{Datasets}

\textbf{Office-Home}~\cite{b14} contains around 15,500 images. It comprises four distinct domains: Art, Clipart, Product, and Real-world. 12 different transfer tasks can be derived by permuting the four domains for fair analysis and evaluation.

\textbf{Office-31}~\cite{b13} comprises 31 different classes and has three distinct domains: Amazon, Webcam, and Dslr. 
Six different transfer tasks can be generated by permuting the evaluation of the three domains.

\textbf{Bing-Caltech} is a real-world noisy dataset. Bing~\cite{b15} was created by obtaining images through the Bing engine corresponding to each category label present in the Caltech-256 dataset~\cite{b36}. As these data were collected on Bing, they are inevitably subject to label noise. We use Bing as the noisy source domain and Caltech as the target domain.

Both Office-Home and Office-31 are inherently clean. We introduce label noise to these datasets to simulate the scenario of domain adaptation under noisy environments. Specifically, if there are a total of $C$ classes, for each class, there will be a probability of $p_{noise}$ for the label to be flipped to one of the other $C-1$ classes with equal probability, where $p_{noise}$ is the noisy rate, ranging from 0 to 1.

\subsection{Comparison Methods}

We compare our methods, Confusing Pair Correction (CPC) and its variant CPC (CAN) based on CAN~\cite{can}, with a series of state-of-the-art methods for unsupervised domain adaptation under noisy environments, such as TCL~\cite{tcl}, RDA~\cite{rda}, CSR~\cite{csr} and UMRDA~\cite{b33}. We also compare our method CPC with state-of-the-art deep learning, curriculum learning, and UDA methods such as SPL~\cite{spl}, MentorNet~\cite{mentornet}, DAN~\cite{dan}, RTN~\cite{rtn}, DANN~\cite{dann}, ADDA~\cite{adda}, and MDD~\cite{mdd}.

\subsection{Results}
The results on Office-Home with 40\% label noise are shown in Table~\ref{result65}. The results on Office-31 under 40\% label noise and on Bing-Caltech are presented in Table~\ref{result31} and Table~\ref{resultbc}.

From Table~\ref{result65}, we observe that our method outperforms all comparison methods on average. 
Our method CPC achieves an average accuracy of 63.5\% on Office-Home, outperforming the previous state-of-the-art method CSR (CAN)~\cite{can} by 4.2\%. In addition, based on CAN, our method CPC (CAN) outperforms CSR (CAN) by 5.4\% on average. Also, CPC (CAN) outperforms CSR (CAN) in 10 out of the 12 total transfer tasks. Especially in the Ar$\rightarrow$Rw task, our accuracy is 76.8\%, surpassing CSR (CAN)'s 65.3\% by 11.5\%. Our method effectively handles label noise, reducing its impact on learning a discriminative model that mitigates the gap between the source and target domain. We also observe that CPC can enhance tasks that already have high accuracy, such as Ar$\rightarrow$Rw. The superiority of our method over existing state-of-the-art methods validates the effectiveness of CPC for domain adaptation under noisy environments. 

From Table~\ref{result31}, our method CPC achieves an average accuracy of 86.3\%, outperforming the RDA \cite{rda} and TCL \cite{tcl}. 
In terms of CAN-based methods, CPC (CAN) achieves an average accuracy of 87.4\%, outperforming CSR (CAN) and UMRDA by 0.6\% and 0.3\% on average, respectively. This is attributed to the robustness of CAN~\cite{can} for addressing label noise, allowing us to further mitigate the adverse impact of label noise.
\begin{figure}[H]
\centering
\includegraphics[width=0.85\columnwidth]{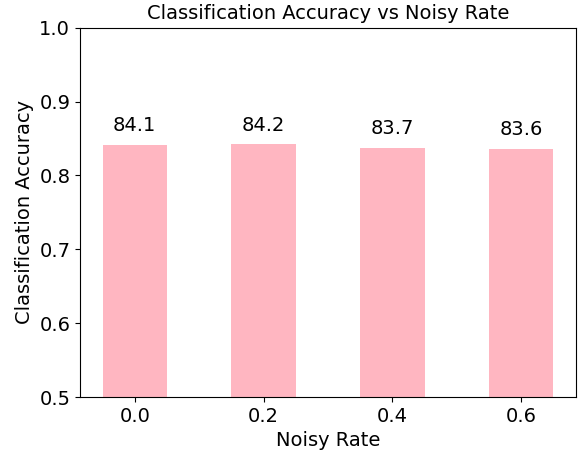}
\caption{The classification accuracy of target domain refinement under different noisy rates in Office-31.}
\label{diffpro}
\end{figure}
\begin{figure}[H]
\centering
\subfigure[puncher]{
    \label{puncher}
    \includegraphics[width=0.2\textwidth]{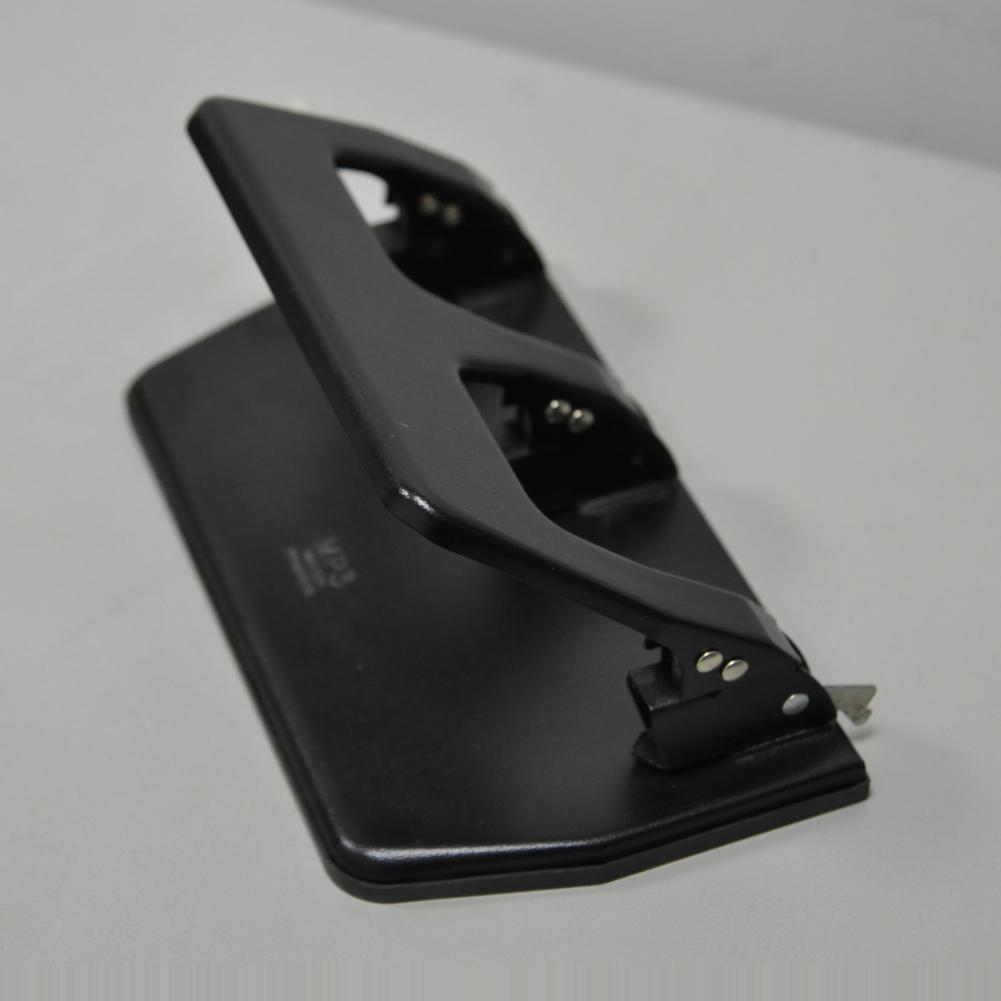}
}
\subfigure[stapler]{
    \label{stapler}
    \includegraphics[width=0.2\textwidth]{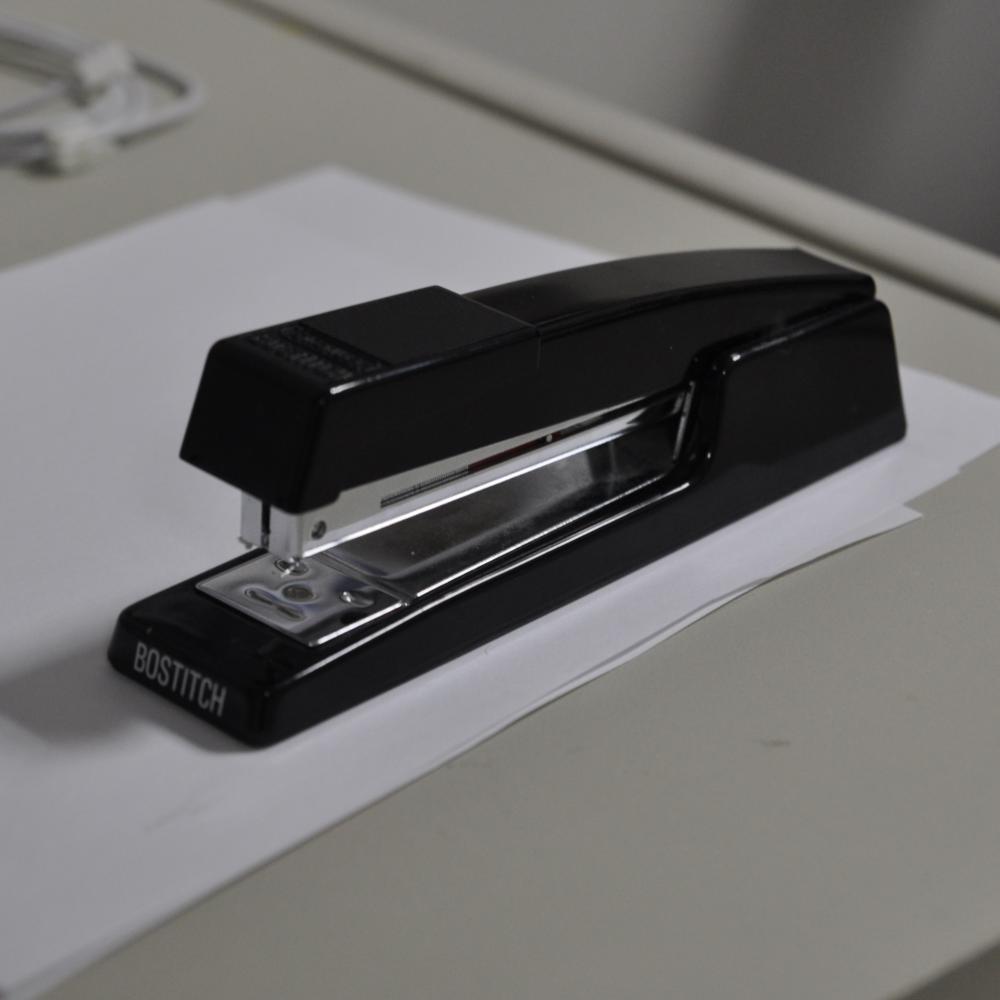}
}
\subfigure[mouse]{
    \label{mouse}
    \includegraphics[width=0.2\textwidth]{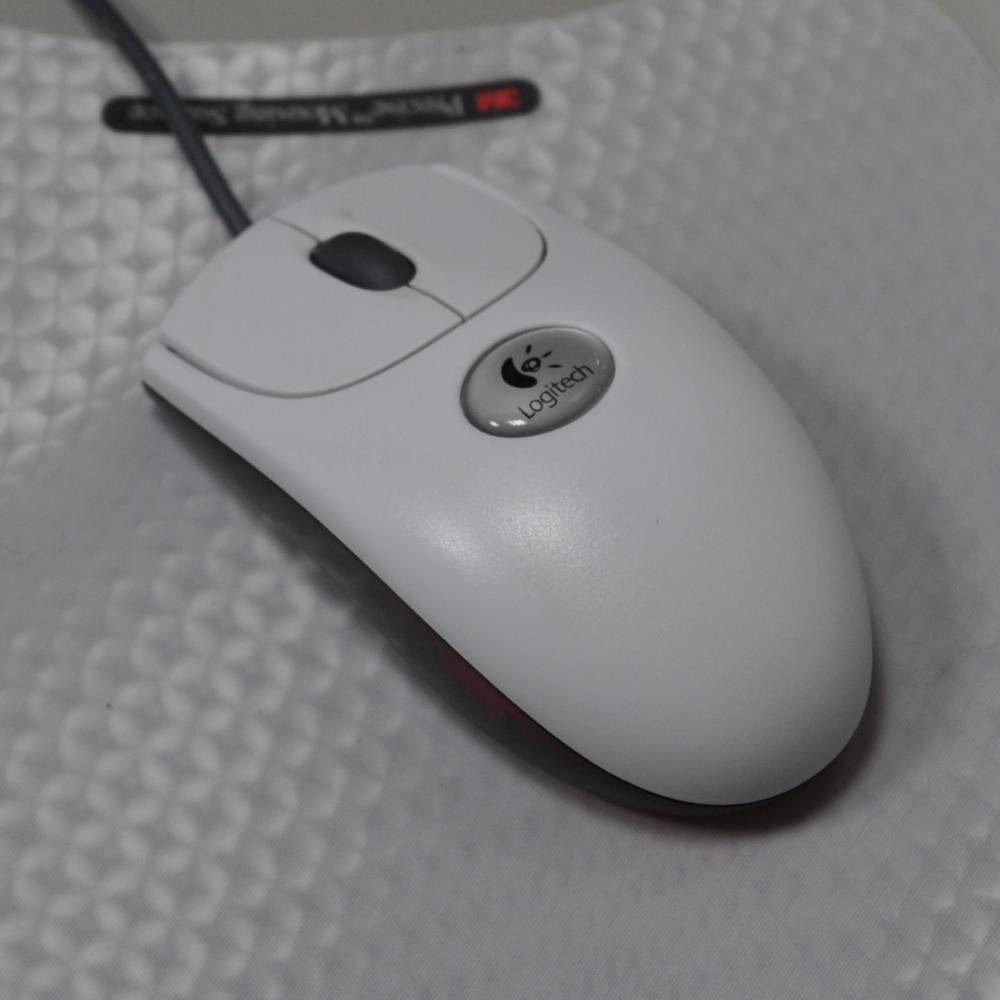}
}
\subfigure[speaker]{
    \label{speaker}
    \includegraphics[width=0.2\textwidth]{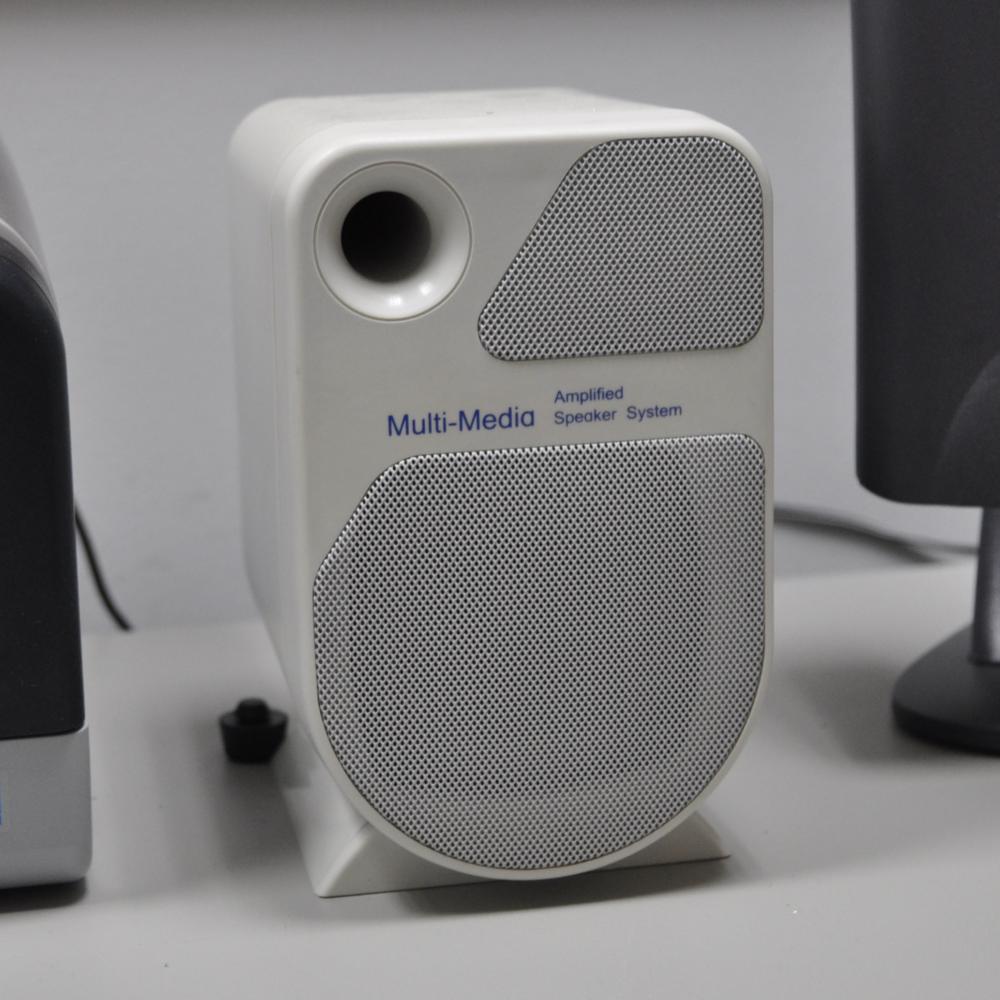}
}
\vspace{-1ex}

\caption{Examples of confusing pairs in domain Dslr. (a) and (b) is a confusing pair. (c) and (d) is another pair.}
\vspace{-2ex}
\label{sampleshow}
\end{figure}
From Table \ref{resultbc}, our approach achieves an accuracy of 85.5\% for the CAN-based method, outperforming RDA, CAN, and CSR (CAN) and getting comparable results with UMRDA~\cite{b33}. The results obtained from this real-world noisy dataset further validate the effectiveness of our model. The reason why UMRDA slightly outperforms our approach is that the uncertainty rank, a component of UMRDA, utilizes two batches of samples for training and then proceeds to generate two models, which involve more intricate calculations than our approach.

\subsection{Ablation Studies}
In this section, we examine the effectiveness of target domain refinement and confusing pair correction separately. Then, we use an example to demonstrate how the label correction method improves the classification accuracy. We also provide further insights into the domain refinement method.


\textbf{Target domain refinement.} To analyze the effectiveness of this module, we conduct experiments using only the target domain refinement for the Office-31 dataset under 40\% label corruption. Recall that Eq. (5) selects $\tau$ samples closest to the prototype within each category, making the classification boundaries clearer. In our experiment, we set $\tau=0.8$.
The experiment consists of two parts: the first one involves the target domain refinement without Eq. (5). The second part involves the target domain refinement with Eq. (5).


The experimental results are shown in Table~\ref{resultaba}. We find that selecting $80\%$ samples closest to the prototype within each category performs better than selecting all samples. Specifically, the accuracy of selecting $80\%$ samples is 83.6\%, 0.3\% higher than selecting all samples. This is because we drop some noisy samples and clarify the classification boundaries. It is worth noting that while the average accuracy of selecting all samples is lower than the one of selecting $80\%$, the accuracy for task D$\rightarrow$W and W$\rightarrow$D is higher. This is because the classification boundaries of different classes in these two tasks are relatively clear. Removing samples will result in the loss of useful information. However, the classification boundaries of different classes in the other four tasks are relatively ambiguous, and discarding boundary samples can remove more noisy samples, leading to clearer classification boundaries. Furthermore, applying the target domain refinement with Eq. (5), our average transfer accuracy is 83.6\%, which is comparable to RDA and outperforms the TCL method by 7.0\%.


\begin{figure}[t]
\centering
\vspace{-2ex}
\includegraphics[width=\columnwidth]{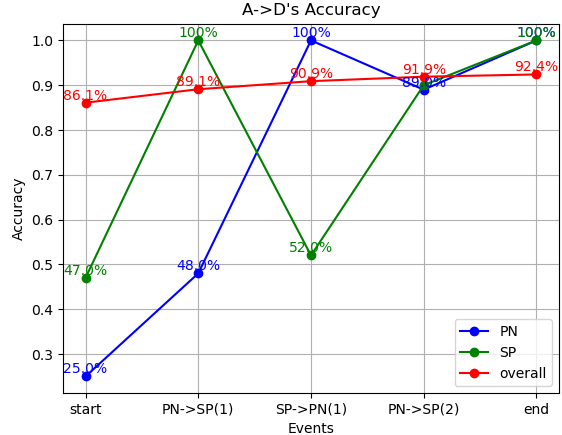}
\caption{Demonstration of how confusing pair correction contributes to improving the accuracy of the confusing pair and overall accuracy. The numbers in parentheses represent the number of times the confusing pair has been corrected.}
\vspace{-2ex}
\label{adacc}
\end{figure}

We also calculate the classification accuracy of the target domain refinement under different noisy rates in the Office-31 dataset, as depicted in Figure \ref{diffpro}. We notice that the accuracy of Office-31 under 60\% label noise is comparable to that without label noise. This result further shows that the target domain refinement is very robust to label noise.

\textbf{Most confusing pair correction.} Based on the target domain refinement, we examine our label correction method. As shown in Table~\ref{resultaba}, the accuracy increases from 83.6\% to 86.3\%, which is a significant improvement of 3.7\%. 

Initially, after the target domain refinement, two classes having similar features will gradually overlap in the high-dimension feature space. The samples of the two classes are both confused and may switch from each other. But as the model converges, one class will become dominant and stable, while the other class will fall into the dominant class. The model first gets the confusing pair (Figure~\ref{sampleshow} shows two confusing pairs in task A$\rightarrow$D). Then, we correct the labels directed by the confusing pairs. Figure \ref{adacc} displays the line chart illustrating how confusing pair correction improves the overall accuracy throughout the training process for task A$\rightarrow$D. From Figure \ref{adacc}, we can observe the impact of confusing pair correction during one of our training process. Initially, the classification accuracy for both classes ``puncher'' (PN) and ``stapler'' (SP) is low. As the model identifies class ``puncher'' and class ``stapler'' as the most confusing pair, it corrects class ``puncher'' in the noisy samples into class ``stapler'', resulting in a 100.0\% accuracy for class ``stapler''. While class ``puncher'' still has an accuracy of only 48.0\%. Subsequently, the model corrects class ``stapler'' back to class ``puncher''. At this point, class ``puncher'' achieves 100.0\% accuracy, while the accuracy of class ``stapler'' drops again. However, during the second correction of class ``puncher'' to class ``stapler'', both classes reach an accuracy of around 90.0\%. Ultimately, class ``puncher'' and class ``stapler'' are correctly classified and the overall classification accuracy reaches 92.4\%. 
Thus, the most confusing pair correction method plays a crucial role in achieving overall accuracy in UDA under label noise.
\begin{figure}[t]
\centering
\includegraphics[width=0.95\columnwidth]{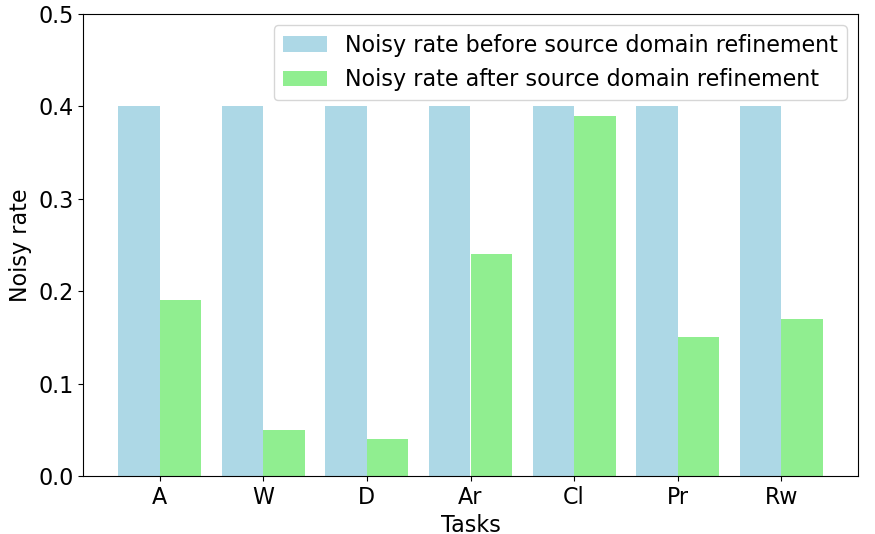}
\caption{The noisy rates of source domain before and after source domain refinement.}
\label{selfprototype}
\end{figure}

\subsection{Further Remark}
\textbf{Source domain refinement} is also a promising method to mitigate the label noise on the source domain. 
Specifically, we use the prototype constructed with a labeled source domain to refine the labels of the source domain. The results are shown in Figure \ref{selfprototype}. We find that refining the labels just once could effectively mitigate a lot of label noise. For example, the noise rate in the domain Dslr notably decreases from 40.0\% to 3.6\% after applying this method. 

Therefore, we can use cleaner source domain data for training. However, we need notice that source domain refinement is not effective on the Clipart domain as it only reduces the label noise by 0.1\%. Therefore, to effectively utilize the source domain refinement, further efforts are required.




\section{Conclusion}
We address a more challenging task, unsupervised domain adaptation under noisy environments.
In this situation, the model can fit noisy labels and perform worse.
To address the problem that the model struggles to classify two similar classes after target domain refinement, we apply the confusing pair correction strategy to clarify the classification boundaries.
We assign pair labels to each sample after aligning the source and target domains. We then identify the most confusing pair among the hard classes and correct the noisy samples of the pair.
Evaluations of three benchmarks show the effectiveness and robustness of our method.
While our method adeptly identifies and corrects certain confusing pairs, it may struggle in more complex scenarios involving three or more similar classes, making handling such cases challenging. Moreover, the model relies on selecting pairs accurately. If the model makes a wrong selection, the model's effectiveness may be compromised.

\section{Acknowledgements}
  This work was supported in part by the National Key R\&D
  Program of China under Grant 2018AAA0102000, and in part
  by National Natural Science Foundation of China: 62022083, 62236008, 61931008 and U21B2038.

\bibliography{main}

\begin{thebibliography}{34}
\providecommand{\natexlab}[1]{#1}

\bibitem[{Bergamo and Torresani(2010)}]{b15}
Bergamo, A.; and Torresani, L. 2010.
\newblock Exploiting weakly-labeled web images to improve object
  classification: a domain adaptation approach.
\newblock \emph{Advances in neural information processing systems}, 23.

\bibitem[{Chen et~al.(2020)Chen, Zhao, Liu, and Cai}]{b31}
Chen, M.; Zhao, S.; Liu, H.; and Cai, D. 2020.
\newblock Adversarial-learned loss for domain adaptation.
\newblock In \emph{Proceedings of the AAAI conference on artificial
  intelligence}, volume~34, 3521--3528.

\bibitem[{Ganin et~al.(2016)Ganin, Ustinova, Ajakan, Germain, Larochelle,
  Laviolette, Marchand, and Lempitsky}]{b22}
Ganin, Y.; Ustinova, E.; Ajakan, H.; Germain, P.; Larochelle, H.; Laviolette,
  F.; Marchand, M.; and Lempitsky, V. 2016.
\newblock Domain-adversarial training of neural networks.
\newblock \emph{The journal of machine learning research}, 17(1): 2096--2030.

\bibitem[{Ganin et~al.(2017)Ganin, Ustinova, Ajakan, Germain, Larochelle,
  Laviolette, Marchand, and Lempitsky}]{dann}
Ganin, Y.; Ustinova, E.; Ajakan, H.; Germain, P.; Larochelle, H.; Laviolette,
  F.; Marchand, M.; and Lempitsky, V. 2017.
\newblock \emph{Domain-Adversarial Training of Neural Networks}, 189–209.

\bibitem[{Goodfellow et~al.(2014)Goodfellow, Pouget-Abadie, Mirza, Xu,
  Warde-Farley, Ozair, Courville, and Bengio}]{b28}
Goodfellow, I.; Pouget-Abadie, J.; Mirza, M.; Xu, B.; Warde-Farley, D.; Ozair,
  S.; Courville, A.; and Bengio, Y. 2014.
\newblock Generative adversarial nets.
\newblock \emph{Advances in neural information processing systems}, 27.

\bibitem[{Griffin, Holub, and Perona(2007)}]{b36}
Griffin, G.; Holub, A.; and Perona, P. 2007.
\newblock Caltech-256 Object Category Dataset.

\bibitem[{Han et~al.(2020)Han, Gui, Cui, and Yin}]{rda}
Han, Z.; Gui, X.-J.; Cui, C.; and Yin, Y. 2020.
\newblock Towards accurate and robust domain adaptation under noisy
  environments.
\newblock \emph{arXiv preprint arXiv:2004.12529}.

\bibitem[{Hoffman et~al.(2018)Hoffman, Tzeng, Park, Zhu, Isola, Saenko, Efros,
  and Darrell}]{b35}
Hoffman, J.; Tzeng, E.; Park, T.; Zhu, J.-Y.; Isola, P.; Saenko, K.; Efros, A.;
  and Darrell, T. 2018.
\newblock Cycada: Cycle-consistent adversarial domain adaptation.
\newblock In \emph{International conference on machine learning}, 1989--1998.
  Pmlr.

\bibitem[{Jiang et~al.(2018)Jiang, Zhou, Leung, Li, and Fei-Fei}]{mentornet}
Jiang, L.; Zhou, Z.; Leung, T.; Li, L.-J.; and Fei-Fei, L. 2018.
\newblock Mentornet: Learning data-driven curriculum for very deep neural
  networks on corrupted labels.
\newblock In \emph{International conference on machine learning}, 2304--2313.
  PMLR.

\bibitem[{Jin et~al.(2020)Jin, Wang, Long, and Wang}]{b23}
Jin, Y.; Wang, X.; Long, M.; and Wang, J. 2020.
\newblock Minimum class confusion for versatile domain adaptation.
\newblock In \emph{Computer Vision--ECCV 2020: 16th European Conference,
  Glasgow, UK, August 23--28, 2020, Proceedings, Part XXI 16}, 464--480.
  Springer.

\bibitem[{Kang et~al.(2019)Kang, Jiang, Yang, and Hauptmann}]{can}
Kang, G.; Jiang, L.; Yang, Y.; and Hauptmann, A.~G. 2019.
\newblock Contrastive adaptation network for unsupervised domain adaptation.
\newblock In \emph{Proceedings of the IEEE/CVF conference on computer vision
  and pattern recognition}, 4893--4902.

\bibitem[{Kumar, Packer, and Koller(2010)}]{spl}
Kumar, M.; Packer, B.; and Koller, D. 2010.
\newblock Self-paced learning for latent variable models.
\newblock \emph{Advances in neural information processing systems}, 23.

\bibitem[{Lee et~al.(2019)Lee, Batra, Baig, and Ulbricht}]{b27}
Lee, C.-Y.; Batra, T.; Baig, M.~H.; and Ulbricht, D. 2019.
\newblock Sliced wasserstein discrepancy for unsupervised domain adaptation.
\newblock In \emph{Proceedings of the IEEE/CVF conference on computer vision
  and pattern recognition}, 10285--10295.

\bibitem[{Long et~al.(2015)Long, Cao, Wang, and Jordan}]{dan}
Long, M.; Cao, Y.; Wang, J.; and Jordan, M. 2015.
\newblock Learning transferable features with deep adaptation networks.
\newblock In \emph{International conference on machine learning}, 97--105.
  PMLR.

\bibitem[{Long et~al.(2018)Long, Cao, Wang, and Jordan}]{b21}
Long, M.; Cao, Z.; Wang, J.; and Jordan, M.~I. 2018.
\newblock Conditional adversarial domain adaptation.
\newblock \emph{Advances in neural information processing systems}, 31.

\bibitem[{Long et~al.(2016)Long, Zhu, Wang, and Jordan}]{rtn}
Long, M.; Zhu, H.; Wang, J.; and Jordan, M.~I. 2016.
\newblock Unsupervised domain adaptation with residual transfer networks.
\newblock \emph{Advances in neural information processing systems}, 29.

\bibitem[{Ma, Zhang, and Xu(2019)}]{b18}
Ma, X.; Zhang, T.; and Xu, C. 2019.
\newblock Deep multi-modality adversarial networks for unsupervised domain
  adaptation.
\newblock \emph{IEEE Transactions on Multimedia}, 21(9): 2419--2431.

\bibitem[{Pan and Yang(2010)}]{b20}
Pan, S.~J.; and Yang, Q. 2010.
\newblock A Survey on Transfer Learning.
\newblock \emph{IEEE Transactions on Knowledge and Data Engineering},
  1345–1359.

\bibitem[{Pan et~al.(2019)Pan, Yao, Li, Wang, Ngo, and Mei}]{b24}
Pan, Y.; Yao, T.; Li, Y.; Wang, Y.; Ngo, C.-W.; and Mei, T. 2019.
\newblock Transferrable prototypical networks for unsupervised domain
  adaptation.
\newblock In \emph{Proceedings of the IEEE/CVF conference on computer vision
  and pattern recognition}, 2239--2247.

\bibitem[{Pei et~al.(2018)Pei, Cao, Long, and Wang}]{b34}
Pei, Z.; Cao, Z.; Long, M.; and Wang, J. 2018.
\newblock Multi-adversarial domain adaptation.
\newblock In \emph{Proceedings of the AAAI conference on artificial
  intelligence}, volume~32.

\bibitem[{Saenko et~al.(2010)Saenko, Kulis, Fritz, and Darrell}]{b13}
Saenko, K.; Kulis, B.; Fritz, M.; and Darrell, T. 2010.
\newblock Adapting visual category models to new domains.
\newblock In \emph{Computer Vision--ECCV 2010: 11th European Conference on
  Computer Vision, Heraklion, Crete, Greece, September 5-11, 2010, Proceedings,
  Part IV 11}, 213--226. Springer.

\bibitem[{Saito et~al.(2018)Saito, Watanabe, Ushiku, and Harada}]{b29}
Saito, K.; Watanabe, K.; Ushiku, Y.; and Harada, T. 2018.
\newblock Maximum classifier discrepancy for unsupervised domain adaptation.
\newblock In \emph{Proceedings of the IEEE conference on computer vision and
  pattern recognition}, 3723--3732.

\bibitem[{Shu et~al.(2019)Shu, Cao, Long, and Wang}]{tcl}
Shu, Y.; Cao, Z.; Long, M.; and Wang, J. 2019.
\newblock Transferable curriculum for weakly-supervised domain adaptation.
\newblock In \emph{Proceedings of the AAAI conference on artificial
  intelligence}, volume~33, 4951--4958.

\bibitem[{Sun and Saenko(2016)}]{b25}
Sun, B.; and Saenko, K. 2016.
\newblock Deep coral: Correlation alignment for deep domain adaptation.
\newblock In \emph{Computer Vision--ECCV 2016 Workshops: Amsterdam, The
  Netherlands, October 8-10 and 15-16, 2016, Proceedings, Part III 14},
  443--450. Springer.

\bibitem[{Tzeng et~al.(2017)Tzeng, Hoffman, Saenko, and Darrell}]{adda}
Tzeng, E.; Hoffman, J.; Saenko, K.; and Darrell, T. 2017.
\newblock Adversarial discriminative domain adaptation.
\newblock In \emph{Proceedings of the IEEE conference on computer vision and
  pattern recognition}, 7167--7176.

\bibitem[{Tzeng et~al.(2014)Tzeng, Hoffman, Zhang, Saenko, and Darrell}]{b26}
Tzeng, E.; Hoffman, J.; Zhang, N.; Saenko, K.; and Darrell, T. 2014.
\newblock Deep domain confusion: Maximizing for domain invariance.
\newblock \emph{arXiv preprint arXiv:1412.3474}.

\bibitem[{Venkateswara et~al.(2017)Venkateswara, Eusebio, Chakraborty, and
  Panchanathan}]{b14}
Venkateswara, H.; Eusebio, J.; Chakraborty, S.; and Panchanathan, S. 2017.
\newblock Deep hashing network for unsupervised domain adaptation.
\newblock In \emph{Proceedings of the IEEE conference on computer vision and
  pattern recognition}, 5018--5027.

\bibitem[{Yan et~al.(2019)Yan, Li, Wang, Li, Xu, and Zuo}]{b19}
Yan, H.; Li, Z.; Wang, Q.; Li, P.; Xu, Y.; and Zuo, W. 2019.
\newblock Weighted and class-specific maximum mean discrepancy for unsupervised
  domain adaptation.
\newblock \emph{IEEE Transactions on Multimedia}, 22(9): 2420--2433.

\bibitem[{Zhang et~al.(2019)Zhang, Liu, Long, and Jordan}]{mdd}
Zhang, Y.; Liu, T.; Long, M.; and Jordan, M. 2019.
\newblock Bridging theory and algorithm for domain adaptation.
\newblock In \emph{International conference on machine learning}, 7404--7413.
  PMLR.

\bibitem[{Zhuo et~al.(2019)Zhuo, Wang, Cui, and Huang}]{uodr}
Zhuo, J.; Wang, S.; Cui, S.; and Huang, Q. 2019.
\newblock Unsupervised open domain recognition by semantic discrepancy
  minimization.
\newblock In \emph{Proceedings of the IEEE/CVF Conference on Computer Vision
  and Pattern Recognition}, 750--759.

\bibitem[{Zhuo, Wang, and Huang(2022)}]{b33}
Zhuo, J.; Wang, S.; and Huang, Q. 2022.
\newblock Uncertainty Modeling for Robust Domain Adaptation Under Noisy
  Environments.
\newblock \emph{IEEE Transactions on Multimedia}, 1--14.

\bibitem[{Zhuo et~al.(2017)Zhuo, Wang, Zhang, and Huang}]{b30}
Zhuo, J.; Wang, S.; Zhang, W.; and Huang, i. 2017.
\newblock Deep Unsupervised Convolutional Domain Adaptation.
\newblock In \emph{Proceedings of the 25th ACM international conference on
  ultimedia}.

\bibitem[{Zhuo et~al.(2023)Zhuo, Zhao, Cui, Huang, and Wang}]{afs}
Zhuo, J.; Zhao, X.; Cui, S.; Huang, Q.; and Wang, S. 2023.
\newblock Adaptive Feature Swapping for Unsupervised Domain Adaptation.
\newblock In \emph{Proceedings of the 31st ACM International Conference on
  Multimedia}, 7017--7028.

\bibitem[{Zuo et~al.(2021)Zuo, Yao, Zhuang, and Xu}]{csr}
Zuo, Y.; Yao, H.; Zhuang, L.; and Xu, C. 2021.
\newblock Seek common ground while reserving differences: A model-agnostic
  module for noisy domain adaptation.
\newblock \emph{IEEE Transactions on Multimedia}, 24: 1020--1030.

\end{thebibliography}

\end{document}